%% file: AISTATS_camera_ready.tex
\RequirePackage[loading]{tracefnt}
\documentclass[twoside]{article}

\usepackage[accepted]{aistats2025}
\usepackage[inkscapelatex=false]{svg}
\usepackage{booktabs} 
\usepackage[pagebackref,breaklinks,colorlinks]{hyperref}
\hypersetup{
    colorlinks=true,
    linkcolor= blue,
    filecolor=violet,      
    urlcolor= blue,
     citecolor  = blue, 
    pdfpagemode=FullScreen,
}
\usepackage[authoryear, sort&compress, round]{natbib}
\usepackage{import}
\subimport{./}{packages}

\subimport{./}{macros}

\usepackage[noabbrev,capitalize,nameinlink]{cleveref}

\begin{document}

%

%
\runningauthor{Y. Cohen, 
A. Navarro, 
J. Frellsen, 
R.E Turner, 
R. Riemer, 
A. Pakman}
 
\twocolumn[

\aistatstitle{Bayesian Circular Regression with  von Mises Quasi-Processes}

\aistatsauthor{
\quad\quad\quad\quad\quad Yarden Cohen${}^{1,2,3}$
\And
\quad\quad\quad \quad Alex Navarro${}^4$
\And
\quad Jes Frellsen${}^5$
\And
Richard E. Turner${}^6$
\And
Raziel Riemer${}^1$
\And
\!\!\!\!\!\!\!\!\!\!\! Ari Pakman${}^{1,2,3}$}

\aistatsaddress{
\qquad \qquad
${}^{1}$Department of Industrial Engineering and Management,
Ben-Gurion University of the Negev
\\
${}^{2}$The School of Brain Sciences and Cognition, Ben-Gurion University of the Negev
\\
${}^{3}$Data Science Research Center, Ben-Gurion University of the Negev
\\
${}^{4}$Unilever
\qquad
${}^{5}$Technical University of Denmark
\qquad
${}^{6}$University of Cambridge
} ]

\begin{abstract}
The need for regression models to predict circular values arises in many scientific fields. In this work we explore a family of expressive and interpretable distributions over circle-valued random functions related to Gaussian processes targeting two Euclidean dimensions conditioned on the unit circle. The  probability model has connections with continuous spin models in statistical physics. 
Moreover, its density is very simple and has maximum-entropy,  unlike previous Gaussian process-based approaches, which use wrapping or radial marginalization. For posterior inference, we introduce a new Stratonovich-like augmentation that lends itself to fast Gibbs  sampling. We argue that transductive learning in these models favors a Bayesian approach to the parameters
and apply our sampling scheme to the Double Metropolis-Hastings algorithm. We present experiments applying this model 
to the prediction of (i) wind directions and (ii) 
 the percentage of the running gait cycle 
 as a function of joint angles. 
\end{abstract}

\section{INTRODUCTION}
Directional or circular data arises in many areas of science, including astronomy, biology, physics, earth science and meteorology.
Since such data is supported on compact manifolds such as tori or (hyper)spheres, it is generally inappropriate  to apply to them standard statistical methods, designed for observations with more familiar supports like $\mathbb{R}^d$.
For reviews of probability models for directional data see~\citet{jupp1989unified,lee2010circular,pewsey2021recent} and for book-length treatments see~\citet{mardia1999directional,jammalamadaka2001topics,pewsey2013circular,ley2017modern,ley2018applied}.

\subimport{.}{transductive}

In this work we explore a model introduced in~\citet{navarro2017multivariate} to perform non-parametric Bayesian regression over circular data.
The model can be obtained by
starting from
a Gaussian process for two-dimensional Euclidean observations
and conditioning on the unit-circle.
 Despite its attractive features
(maximum entropy and a simple distribution),
the model remained practically unexploited
by the lack of effective methods for posterior inference and parameter learning. Here  we address both challenges
by developing novel Markov Chain Monte Carlo (MCMC) techniques
 tailored to this setting, with emphasis on the need for transductive learning
of the model parameters (see~\cref{fig:synth_data}).

The model is reviewed in~\cref{sec:vM},
inference is studied  in~\cref{sec:sampling},
parameter learning  in~\cref{sec:learning}, related works are reviewed in~\cref{sec:related}, followed by  experiments  in~\cref{sec:experiments} and conclusions in
\cref{sec:conclusions}.\footnote{Our code is available at
\url{https://github.com/Yarden231/vMQP}.
}

\section{VON-MISES QUASI-PROCESSES}
\label{sec:vM}
Consider a regression problem where  a set of circular variables $\{\theta_i\}_{i=1}^n$ are observed at  input locations~$\{x_i\}_{i=1}^n$.
Our goal is to predict new values
$\{\phi_i\}_{i=1}^m$ at unseen locations~$\{x^*_i\}_{i=1}^m$. A typical application is in geostatistics, where one wants to predict the direction of wind or ocean waves.
We  denote the combined
unobserved $\bphi$ and observed $\bt$ variables as $\bvphi=[\bphi, \bt]$, define $d=m+n$
and consider a prior density for $\bvphi$ given by
\begin{align}
p(\bvphi|\x)
&\varpropto \exp \left\{
-\frac12
\sum_{i,j=1}^d M_{ij} \cos(\varphi_i-\varphi_j)
\right.
\nn
\\
& \left.
+ \kappa \sum_{i=1}^d  \cos(\varphi_i-\nu)
\right\}.
\label{eq:prior}
\end{align}
Here $\nu \in [-\pi, +\pi]$,
and we restrict to $\kappa >0$ since the
density is invariant under $(\kappa, \nu) \leftrightarrow
(-\kappa, \nu-\pi)$.
The matrix in (\ref{eq:prior})  is the
inverse~$M_{ij}= (K^{-1})_{ij}$
of $K_{ij}= k(x_i,x_j)$, where $k(\cdot,\cdot)$ is a function
that plays a role similar to the covariance in Gaussian processes in Euclidean target space. We call the model~(\ref{eq:prior}) a {\it von Mises Quasi-Process (vMQP)}
to stress the connection with Gaussian processes, although this
is not a stochastic process, as explained in~\cref{sec:learning}.

The posterior distribution
(up to a constant factor)
of the unobserved~$\bphi$, obtained from~(\ref{eq:prior}) by setting constant~$\bt$, is
\begin{align}
& p(\bphi|\bt, \x)
\varpropto
\exp \left\{
\bm{\rho}_c \cdot \cos(\bphi)
+  \bm{\rho}_s \cdot \sin(\bphi)
\right.
\label{eq:conditional_density}
\\
 &
\left.
-\frac12 \cos(\bphi)^\top
\bm{Q}
\cos(\bphi)
-\frac12 \sin(\bphi)^\top
\bm{Q}
\sin(\bphi)
\right\} \,,
\nn
\end{align}
where we defined
\begin{align}
\bm{\rho}_c & = -\bm{M}_{\bphi \bt}  \cos(\bt)
+ \kappa \cos(\nu) \bm{1}_{m},
\\
\bm{\rho}_s  &= -\bm{M}_{\bphi \bt}  \sin(\bt)
+ \kappa \sin(\nu) \bm{1}_{m},
\\
\bm{Q} & = \bm{M}_{\bphi \bphi},
\end{align}
with  $\bm{1}_{m} \in \mathbb{R}^m$ a vector of $1$'s, and
the matrix
\begin{align}
\bm{M}  &\equiv
\begin{bmatrix}
\bm{M}_{\bphi \bphi} & \bm{M}_{\bphi \bt}
\\
\bm{M}_{\bt \bphi}   & \bm{M}_{\bt \bt}
\end{bmatrix}  \in \mathbb{R}^{d \times d}
\end{align}
has entries $M_{ij}$. In the machine learning
literature,  similar distributions
were studied in~\citet{navarro2017multivariate},
which showed that these are maximum entropy distribution with fixed first circular moments and proposed their use for circular regression.
But the practical exploitation of these models has been hindered by the lack of effective methods to
(i) sample efficiently, and (ii) learn the
parameters $w \equiv (\kappa, \nu, \textrm{Kernel parameters})$.
We address both of these challenges in~\cref{sec:sampling,sec:learning}, respectively.

\subsection{Relation to Gaussian Processes}
The prior (\ref{eq:prior}) can be obtained
from a Gaussian Process (GP)  over $d$ 2D Euclidean random function values~$(f_{1,i},f_{2,i}) \in \mathbb{R}^2$, where $f_{1,i}$ and $f_{2,j}$ are uncorrelated for all $i,j$ and share the same
covariance matrix
$K_{ij}= k(x_i,x_j)$.
The density~is
\begin{align}
p(\f_1, \f_2|\x)
&\varpropto \exp \left\{
-\frac12
(\f_1-\bm{\mu}_1)^\top
\bm{M}
(\f_1-\bm{\mu}_1)
\right.
\nn
\\
&
\left.
 -\frac12
(\f_2-\bm{\mu}_2)^\top
\bm{M}
(\f_2 -\bm{\mu}_2)
\right\}.
\label{eq:2d_Gaussian}
\end{align}
Expressing $(f_{1,i}, f_{2,i}) = (r_i \cos(\varphi_i),
r_i \sin(\varphi_i))$,
conditioning on~$r_i=1$ and setting
 $\bm{\mu}_i = {\mu}_i \bm{1}_d$ for $i=1,2$ with $\bm{1}_d$ a vector of $1$'s,  yields (\ref{eq:prior}),
where $\kappa$ and $\nu$ are  functions of ${\mu}_1, {\mu}_2$ and~$\bm{M}$.
In Euclidean space GPs it is common to set
$\mu_1=\mu_2=0$ and subtract from the data the empirical average. In our case, we keep
$\kappa >0$ in general, since $\kappa =0$ implies a uniform distribution for the marginal of each variable $\varphi_i$, an assumption that needs to be evaluated in each case.

\subsection{Including noisy observations}
For noisy observations, we start with the prior~(\ref{eq:prior}), and instead of identifying
the $n$ last components of~$\bvphi$
with the observations $\theta_i$,
we keep them as separate variables and
assume that each observation $\theta_i$ has a 1D von Mises likelihood centered on $\varphi_{m+i}$, leading to a posterior
\begin{align}
p(\bvphi|\bt, \x)
\,
\varpropto  \exp\{ \chi \sum_{i=1}^n  \cos(\theta_i -\varphi_{m+i}) \}
p(\bvphi|\x)
\label{eq:noisy}
\end{align}
where $\chi$ is a parameter.
The equations in the next Sections are presented for the noiseless case, but they are easily extended to the case (\ref{eq:noisy}).

\section{SAMPLING CIRCULAR VARIABLES}
\label{sec:sampling}
Unlike Gaussian processes, closed-form expressions for the posterior mean and variance of the distribution (\ref{eq:conditional_density}) are not available,
thus the need for an efficient sampling approach.
Note that this is not a big limitation, since
even in standard GPs, many quantities of interest, such as expectations of nonlinear
functionals, require posterior samples for their estimation~\citep{wilson2020efficiently,wilson2021pathwise}.

Our method starts by noting that  $\bm{Q} \in \mathbb{R}^{m \times m}$ in (\ref{eq:conditional_density}) is positive definite, since it is a submatrix of $\M$.
Let $\lambda~\in~\mathbb{R}$ be a number bigger than all the eigenvalues
of~$\bm{Q}$. We perform now a Cholesky decomposition,
\begin{equation}
    \A^\top \A = \lambda \bm{I}_{m}  - \bm{Q},
    \label{eq:Cholesky}
\end{equation}
where the r.h.s is positive definite.
The key idea is to augment (\ref{eq:conditional_density}) with a pair of Gaussian random variables $\z_1, \z_2 \in \mathbb{R}^{m}$ with densities
\begin{align}
    p(\z_1|\bphi) &= {\cal N}
    \left(\z_1; \A \cos(\bphi),  \bm{I}_{m} \right)
    \label{eq:Gauss_densities1}
    \\
    p(\z_2|\bphi) & = {\cal N}
    \left(\z_2; \A \sin(\bphi),  \bm{I}_{m} \right).
\label{eq:Gauss_densities2}
\end{align}
Multiplying (\ref{eq:conditional_density}) by both densities in (\ref{eq:Gauss_densities1})-(\ref{eq:Gauss_densities2})  we get the augmented distribution
\begin{align}
    p(\z,\bphi) & = p(\bphi) p(\z_1|\bphi)  p(\z_2|\bphi)
    \nn
    \\
    &     \varpropto
    \exp \left\{
    (\bm{\rho}_c^{\top} + \z_1^{\top} \A)  \cos(\bphi)
    \right.
    \nn
\\
& \left.
    +
    (\bm{\rho}_s^{\top} + \z_2^{\top} \A)  \sin(\bphi)
    -\frac12 \z^{\top} \z     \right\} ,
    \label{eq:joint}
\end{align}
where we denoted $\z = [\z_1, \z_2]$
and omitted the conditioning variables $(\bt, \x)$ to simplify the notation.
Importantly, the terms quadratic in $[\cos(\bphi),\sin(\bphi)]$ in the exponent of
(\ref{eq:conditional_density}) have
canceled in (\ref{eq:joint}).
This linearization of the trigonometric dependence
in the exponent is similar to the Hubbard–Stratonovich transformation in field theory~\citep{altland2010condensed}, and in
machine learning  it has been
similarly applied to interacting log-quadratic binary variables
in~\cite{martens2010parallelizable, zhang2012continuous,ostmeyer2021ising}.

Note that sampling from (\ref{eq:joint}) and keeping only the $\bphi$ samples yields samples from (\ref{eq:conditional_density}).
The advantage of the augmented form  is
that it lends itself to using a simple Gibbs sampler that alternates between
\begin{enumerate}
    \item Sample $\z|\bphi$ from (\ref{eq:Gauss_densities1})-(\ref{eq:Gauss_densities2})
    by
sampling $[{\bm \varepsilon}_1, {\bm \varepsilon}_2] \sim    {\cal N}   \left( 0, \bm{I}_{2m} \right)$
and setting
    \begin{equation}
        \begin{bmatrix}
        \z_1
        \\
        \z_2
        \end{bmatrix}
        =
\begin{bmatrix}
\A \cos(\bphi)
\\
\A \sin(\bphi)
\end{bmatrix}  +
\begin{bmatrix}
{\bm \varepsilon}_1
\\
{\bm \varepsilon}_2
\end{bmatrix}
\label{eq:z_given_phi}
\end{equation}
    \item Sample $\bphi|\z$ from
\begin{equation}
    p(\bphi|\z) \varpropto \prod_{i=1}^m \exp(a_i \cos(\phi_i -\gamma_i) ) \,,
\label{eq:product_vMises}
\end{equation}
where we defined
\begin{align}
    a_i &= \sqrt{b_{c,i}^2 + b_{s,i}^2} \,
\,\,\,\,\,\, \,\,\,    \tan (\gamma_i)  = \frac{b_{s,i}}{b_{c,i}}  \,,
\end{align}
and $b_{c,i}$ and $b_{s,i}$ are the components of
\begin{align}
\bm{b}_{c} & = \bm{\rho}_c +  \A^{\top} \z_1 \,,
\\
\bm{b}_{s} &= \bm{\rho}_s +
\A^{\top} \z_2 \,.
\label{eq:kappas}
\end{align}
\end{enumerate}
The distribution (\ref{eq:product_vMises}) is a product of independent one-dimensional
von Mises distributions,
and can be sampled efficiently using rejection-sampling~\citep{best1979efficient}, as implemented
in standard packages, or using exact Hamiltonian Monte Carlo~\citep{pakman2025super}.

\paragraph{The role of $\lambda$.}
Combining (\ref{eq:z_given_phi}) with (\ref{eq:kappas})
we get
\begin{align}
\bm{b}_{c} = \bm{\rho}_c +
\cos(\bphi) [ \lambda \bm{I}_{m}  - \bm{Q} ]
+ \A^{\top}{\bm \varepsilon}_1
\end{align}
and similarly for $\bm{b}_{s}$. In the $\lambda \rightarrow \infty$ limit,
in iteration~$t$,
we get $\bm{b}_{c} \simeq  \lambda \cos(\bphi^{(t)})$ and the means of $p(\bphi^{(t+1)}|\z)$ in~(\ref{eq:product_vMises})
become $\gamma_i \rightarrow \phi_i^{(t)}$,
hurting the exploration of the circular space.
This non-rigurous heuristic argument suggest $\lambda$ should not be bigger than necessary.
We verify empirically this claim in~\cref{sec:explore_lambda}.


\section{LEARNING THE PARAMETERS} 
\label{sec:learning}
Our learning objective is determined by the fact that, unlike standard Gaussian~\citep{williams2006gaussian} or 
$t$-processes~\citep{shah2014student}, 
 von Mises Quasi-Processes are not consistent under 
 marginalization. Recall that we observe  $\bt$ at locations $\x$ and are interested in 
predictions $\bphi$ at~$\x^*$. Let us express the 
normalized distribution~(\ref{eq:prior}) as
\begin{align}
    p(\bphi, \bt|\x, \x^*,w)  = \frac{e^{-U(\bvphi|w)}}{Z[w]}    \qquad \bvphi = [\bphi, \bt]
\end{align}
where 
\begin{align}
U(\bvphi|w) = &
\frac12 
\sum_{i,j=1}^d M_{w,ij} \cos(\varphi_i-\varphi_j) 
\\
\nn 
& - \kappa \sum_{i=1}^d  \cos(\varphi_i-\nu), 
\\
Z[w]  = &  \int d \bvphi e^{-U(\bvphi|w)},
\label{eq:Z_w_definition}
\end{align}
and $w$ are the parameters we want to learn. 
Since~$\bphi$ are unobserved, 
maximum likelihood corresponds to 
\begin{equation}
\hat{w} =  \argmax_{w} \, \log \int d \bphi \, p(\bphi, \bt|\x, \x^*,w)     .
\label{eq:objective}
\end{equation}
In standard Gaussian or $t$-processes, 
the above expression would reduce to
\begin{equation}
\hat{w} =  \argmax_{w} \, \log p(\bt|\x,w)     ,
\end{equation}
i.e., the test locations $\x^*$ disappear upon marginalization of $\bphi$. But this is not the case in our model,\footnote{Note that since models with different number of prediction locations are not related via marginalization, the Kolmogorov extension theorem~\citep{durrett2019probability} does not apply and this prevents us from calling these models~{\it processes}. 
 }  
and therefore, we must fix the test locations~$\x^*$ at training time, 
a setting known as {\it transductive learning}~\citep{vapnik200624} and illustrated in~\cref{fig:synth_data}.

\subsection{Problems with point estimates}
\label{sec:point_estimates}
The  gradient of 
the learning objective~(\ref{eq:objective}) is
\begin{align}
& \nabla_{w} \log \int d \bphi \, p(\bphi, \bt|\x, \x^*,w) =
\label{eq:cd_gradient}    
\\
&\mathbb{E}_{p_2(\bphi',\bt')}  [\nabla_w U(\bphi',\bt'|w)]
-\mathbb{E}_{p_1(\bphi')} 
[\nabla_w U(\bphi',\bt|w)]
\nn
 \end{align}
where we defined 
\begin{align}
    p_2(\bphi',\bt') &= p(\bphi',\bt'|\x, \x^*,w),
    \\ 
    p_1(\bphi') & = p(\bphi'|\bt,\x, \x^*,w). 
\end{align}
This is the standard contrastive divergence gradient in the presence of latent variables, similar to that used to train 
Restricted Boltzmann Machines~(RBMs)~\citep{carreira2005contrastive}. 
But, unlike the latter case, we have found in experiments, using MCMC estimates for the expectations in~(\ref{eq:cd_gradient}), that this approach is utterly ineffective, arguably due to two differences. First, unlike RBMs, 
we only  learn a small number of parameters (e.g. just three or four parameters in the examples 
in~\cref{sec:experiments}), and thus small amounts of noise or bias in the gradient estimation have potentially deleterious effects. Second, again unlike RBMs, the number of latent variables $m$ in our case can be much higher than the number of observed variables $n$, and thus 
capturing the difference between the two terms in (\ref{eq:cd_gradient}) might require  impractically large numbers of Monte Carlo samples. 
\cref{fig:cont_div} illustrates this problem.

\subimport{.}{cont_div_histograms}

\subsection{A fully Bayesian approach}
To avoid the problems mentioned above, we resort instead to a fully Bayesian approach. We are  interested in sampling from the joint posterior of parameters and unobserved variables, 
\begin{align}
    p(\bphi,w|\bt)  \varpropto 
    p(\bphi, \bt  |w) p(w) ,
\end{align}
where $p(w)$ is a prior distribution and
we omit from now on the conditioning locations $(\x,\x^*)$. 
Using block Gibbs sampling, 
the conditional 
$p(\bphi|\bt, w)$  is sampled 
 with the method of~\cref{sec:sampling}. 
On the other hand, sampling from the conditional 
\begin{align}
p(w|\bvphi)
\varpropto & p(w) \frac{
f(\bvphi|w)}
{Z[w]}, \qquad \bvphi = [\bphi, \bt],
\label{eq:conditional_w}
\end{align}
where we defined
\begin{align}
f(\bvphi|w) = &
\exp \left[  -U(\bvphi|w) \right],
\end{align}
is challenging, because we  lack a 
closed-form expression for the 
normalization constant~$Z[w]$, defined in~(\ref{eq:Z_w_definition}).  Several algorithms have been developed to tackle 
this problem (see~\citet{park2018bayesian} for a review). In the following, we review the 
Exchange, Double Metropolis-Hastings and Bridging algorithms. 

\paragraph{The Exchange and Double MH algorithms.} The Exchange algorithm~\citep{murray2006mcmc}, inspired by~\cite{moller2006efficient}, starts by augmenting~(\ref{eq:conditional_w}) with a freely-chosen proposal distribution~$q(w'|w)$ over new parameters $w'$ and a fictitious data point~$\bxi \in [0,2\pi]^d$ generated by $w'$ on the full space (observed and unobserved), 
\begin{align}
    p(w,w',\bxi| \bvphi)  &= p(w|\bvphi) q(w'|w)
\frac{
f(\bxi|w')}
{Z[w']} \,,
\label{eq:augmented0}
\\
& \varpropto
p(w) q(w'|w)
\frac{
f(\bvphi|w)}
{Z[w]} 
\frac{
f(\bxi|w')}
{Z[w']} \,.
\label{eq:augmented}
\end{align}
One now samples from this joint distribution  by alternating between two Monte Carlo moves:
\begin{enumerate}
    \item 
    Sample 
    \begin{align}
        w' &\sim q(w'|w) \,,
        \\
        \bxi &\sim     f(\bxi|w')/Z[w'] \,.
    \end{align}  
    \item Propose to exchange $w \leftrightarrow w'$ and accept with Metropolis-Hastings (MH) probability
    \begin{align}
&    \min\left(1, \frac{p(w', w, \bxi| \bvphi)}{p(w, w',\bxi| \bvphi)} \right) =      
\nn
 \\   
    &\min\left(1, \frac{p(w') q(w|w') f(\bvphi|w') f(\bxi|w)} {p(w) q(w'|w)  f(\bvphi|w) f(\bxi|w')} \right), 
\label{eq:double_MH}
\end{align}
    where, remarkably, $Z[w]$ and $Z[w']$ cancel. 
\end{enumerate}
If exact sampling of $\bxi$ in Step 1 is difficult, as in our model, it was suggested in~\cite{liang2010double}  
to approximate an exact sample
by running long enough a Markov chain targeting $p(\bxi) \varpropto f(\bxi|w')$, an approach dubbed Double Metropolis-Hastings. 
In our case, we run this Markov chain using an augmentation as in~\cref{sec:sampling}, based on a Cholesky decomposition 
\begin{align}
    \A^\top_{w'}  \A_{w'} &= \bm{I}_{d} \lambda_{w'} - \bm{M}_{w'} ,
    \label{eq:cholesky_full}
\end{align}
where $\lambda_{w'}$ is chosen to make the r.h.s.~positive definite.

\paragraph{Efficient Bridging for the vMQP.} 
Note that if we knew  the normalization constant $Z[w]$ in (\ref{eq:conditional_w}), we could  sample a proposal from $q(w'|w)$ and accept it with MH probability 
    \begin{equation}
    \min\left(1, \frac{p(w') q(w|w') f(\bvphi|w') Z[w]}
    {p(w) q(w'|w)  f(\bvphi|w) Z[w']} \right).    
    \label{eq:exact_MH}
    \end{equation}
Equation (\ref{eq:double_MH}) in the Exchange algorithm corresponds instead to replacing the ratio of  normalizing constants in (\ref{eq:exact_MH}) with  a one-sample importance sampling approximation 
\begin{align}
    \frac{Z[w]}{Z[w']} \simeq \frac{f(\bxi|w)} {f(\bxi|w')}. 
    \label{eq:ratio_normalizing}
\end{align}
where 
\begin{align}
\qquad  \bxi \sim     f(\bxi|w')/Z[w'] \,.    
\end{align}
Now, since it is known that the acceptance rate (\ref{eq:exact_MH}) is maximal~\citep{peskun1973optimum,tierney1998note}, one can expect to increase the acceptance rate (\ref{eq:double_MH}) by using an 
estimate with lower variance than (\ref{eq:ratio_normalizing}). 
The Bridging algorithm~\citep{murray2006mcmc} achieves this 
by further augmenting (\ref{eq:augmented0})-(\ref{eq:augmented})
with $K$ additional variables. 
In~\cref{app:bridging} we review the Bridging algorithm and present an efficient extension tailored to our augmented model (\ref{eq:joint}). 

\section{RELATED WORKS}
\label{sec:related}

\paragraph{Relation to statistical physics.}
Distributions of the type (\ref{eq:conditional_density}) are known in statistical physics as XY- or $O(2)$-models, with the angles $\phi_i$ being a continuous generalization of the $\{\pm1\}$ spins  of Ising models. The case when the $\phi_i$'s are located 
in a $d$-dimensional regular lattice and~$\bm{Q}$ has a sparse structure, with non-zero entries only between  nearest-neighbours, has been intensely researched since the 1960s~\citep{friedli_velenik_2017}, in particular for~$d=2$. Although physicists have developed 
several specialized algorithms to sample from XY-models, their efficiency depends on sign or sparsity properties of the~$\bm{Q}$ matrix 
absent in  our Bayesian regression setting, characterized by unsigned, dense $\bm{Q}$ matrices.
For example, cluster flipping  algorithms perform  well in the 2D lattice ferromagnetic regime~\citep{wolff1989collective} 
(when non-zero entries of~$\bm{Q}$ are negative), but fail for spin-glasses~\citep{kessler1990unbridled} (when non-zero entries of $\bm{Q}$ have both signs), the relevant case for us.
Worm algorithms~\citep{prokof2001worm,wang2005worm}
rely on the lattice nearest-neighbour topology, 
absent in our case with dense $\bm{Q}$s. 
Finally, a piecewise-deterministic Monte Carlo sampler 
for the XY-model~\citep{michel2015event,manon2016irreversible}, with exactly-solvable event times, 
 is inefficient for non-sparse~$\bm{Q}$ since the times between consecutive events tend to zero as the number of non-zero elements in $\bm{Q}$ grows. 

\paragraph{Other circular models from Gaussian processes.}
Gaussian process-like models for this task have been obtained in the past by starting from a GP in an Euclidean target space and applying a transformation to yield a distribution in circular space. 
The most popular approaches are wrapping and projecting~\citep{jona2018spatial}. 
Wrapping  consists in imposing an equivalence structure on the target space, namely 
$$y_i \simeq y_i + 2\pi k,$$ for $k \in \mathbb{Z}$ for all $i$. The projecting approach considers a two-dimensional space similar 
to~(\ref{eq:2d_Gaussian}) and, after
expressing the target in polar coordinates~$(f_{1,i}, f_{2,i}) = (r_i \cos(\varphi_i),
r_i \sin(\varphi_i))$, it marginalizes the radial components $r_i$. 
Both approaches have been extensively explored in the literature. In both cases, the resulting probability densities lack a 
 simple form such as~(\ref{eq:prior}), 
but the normalization constant is known since it is inherited from the original GP.

Wrapped distributions  require approximations due to infinite sums implicit in their definition and are often truncated at the third harmonic~\citep{mardia1999directional}. More recently, adaptive truncation schemes~\citep{jona2012spatial} and modelling the truncation point as a latent variable~\citep{jona2014models} were suggested to alleviate the  errors caused by truncation. Fully Bayesian approaches, using
variations of Gibbs samplers for learning and inference have been 
developed both for the wrapped approach in~\cite{ferrari2009wrapping, jona2018spatial,jona2020circspacetime, marques2022wrapped},
and for the projecting approach in~\cite{nunezantonio2005bayesian, nunezantonio2011bayesian,nunezantonio2014bayesian,  hernandezstumpfhauser2017general,
 jona2018spatial,jona2020circspacetime,zito2023projected}.

 \paragraph{Variational inference.}
 Despite the attractive simplicity of the density~(\ref{eq:prior}), this model seems to have remained unexplored since the work by~\cite{navarro2017multivariate}, which studied an approximation to the posterior (\ref{eq:conditional_density}) via variational inference, with a proposal given by a product of one-dimensional von Mises distributions.
 But such an approach is inefficient, because 
 the resulting KL divergences lack closed form and 
require expensive numerical evaluations.

\section{EXPERIMENTS}
\label{sec:experiments}

\subimport{.}{ess_lambda}

\subsection[]{Selecting the $\lambda$ parameter}
\label{sec:explore_lambda}
As mentioned in~\cref{sec:sampling}, we expect better mixing of our augmented Gibbs sampler for smaller values of the parameter $\lambda$ in~(\ref{eq:Cholesky}).  
In~\cref{fig:ess_lambda}, we verify this claim for the 
the data presented in~\cref{fig:synth_data} with $m=10$. 


\subsection{Sampler comparisons}
In~\cref{fig:sampler_performace}  we compare the efficiency of Gibbs sampling 
and Hamiltonian Monte Carlo (HMC)~\citep{neal2011mcmc} applied to 
both the non-augmented (\ref{eq:conditional_density}) and augmented~(\ref{eq:joint}) distributions for the same data 
 as in~\cref{fig:ess_lambda}. 
The Gibbs conditionals in 
(\ref{eq:conditional_density}) are 1D von Mises. 
In some cases (notably for large $m$), we found that it is possible to adjust the parameters of HMC (size and number of steps in each iteration) to be more efficient than the augmented Gibbs sampler, but the latter offers a simple and straightforward approach without the need to tune parameters.


\subimport{.}{sampler_performance}

\subsection{Wind directions in Germany}
\label{sec:wind_predictions}
\subimport{./}{wind_test_table} 
\subimport{./}{wind_directions}

In this experiment, we considered the problem of predicting wind directions at selected locations
based on spatial proximity. 
We used data publicly available on the website of 
the German weather service~\cite{dwd},
which consists of measurements collected at 260 weather stations every 10 minutes.
We considered a single observation from the final day of the calm weather period studied in~\cite{marques2022wrapped}. We only considered the spatial location as a the covariate. We note that our model is admittedly too simple,
and other 
variables such as humidity, temperature, altitude, etc. 
should be used in more realistic models. 
For our vMQP model we considered both Gaussian and Exponential covariance kernels of the forms, respectively, 
\begin{align}
    K_{\text{G}}(\x_{i}, \x_j) &= \sigma^2 \exp\left( -\frac{|| \x_{i} -\x_{j}||^2 }{2l^2} \right),    
\label{eq:gaussian_kernel}
\\
K_{\text{E}}(\x_{i}, \x_j) &= \sigma^2 \exp\left( -\frac{|| \x_{i} -\x_{j}|| }{l} \right),    
\label{eq:exponential_kernel}
\end{align}
for $i,j = 1\ldots 260$, 
with two parameters $(\sigma^2,l)$.
Here  $\x_{i}$ is the 2D spatial coordinates (longitude and latitude) of each weather station.

We compared vMQP predictions with both wrapped and projected GPs, using the implementations of the~\texttt{CircSpaceTime}~R 
package~\citep{jona2020circspacetime}. Wrapped and Projected GP models were only tested with the Exponential kernel due to numerical instability with the Gaussian kernel. Train/test splitting involved test sizes of 10\%, 20\%, 30\% and 40\% of the data. 
\cref{fig:wind_direction_variance} (Left) shows the training and test wind directions and the circular mean of the vMQP predictions for the 20\% test size, while \cref{fig:wind_direction_variance} (Right)
shows predicted circular variances~(see \cref{app:definitions} for definitions).

All the predictive distributions were evaluated against the test data using the circular continuous ranked probability score (CRPS)~\citep{grimit2006continuous} (see~\cref{app:crps} for the definition). 
The results, presented in ~\cref{tab:wind}, report the mean and standard deviation of CRPS across seven random train/test splits. The results confirm the advantage of the vMQP model and show that the Exponential kernel is better in most cases. See~\cref{app:wind_experiment} for more details.

\subimport{.}{gait}

 \subsection{Percentage of running gait cycle from joints angles}
\label{sec:percentage_prediction}
When a human runs, the positions of lower-limb joints
go through a recurrent trajectory known as the {\it gait cycle}, illustrated in~\cref{fig:gait}.
The joints' positions along the cycle are described by 
three joint angles, which are critical for understanding the neuromechanics and energetics of human locomotion~\citep{winter1983energy}. 
In applications, joint angles and 
the phases in the running cycle are used to guide control software in exoskeletons~\citep{gad2022biomechanical} and controllers for commercial 
devices~\citep{sanz2014generation,esquenazi2012rewalk}
and prostheses~\citep{markowitz2011speed,sup2008design}.

\subimport{.}{joints_test}

In this experiment, we consider the task of predicting 
the phase in the running gait cycle, measured as 
a percentage $t \in (0,100]$, as a function of the joint angles. Since $t$ is circular,
the vMQP provides an appropriate prediction model. 
We used data  from~\cite{shkedy2022parametric}, 
who collected data from 16 healthy adults,  while running on treadmill at several surface 
gradients.\footnote{The surface gradient is  $100 \, \times$ tangent of inclination angle, and is indicated with a $\%$ symbol.} See~\cref{app:gait_experiment} for more details. 
The data consists of values of the three angles defined in~\cref{fig:gait}, estimated at 100 values of the percentage $t \in \{1,2 \ldots 100\}$ 
from individuals  running at five surface gradients $s \in \{ 0 \%, \pm 5 \%, \pm 10 \% \}$. We used four gradients $s \in \{ \pm 5 \%,\pm 10  \%\}$ for training, yielding 400 training points. 
Training data for $s = \pm 10 \% $ are shown in~\cref{fig:gait_data}.  
Testing was performed on triplets of angles measured 
at surface gradient~$s=0 \%$ at 20 points uniformly selected along the gait cycle.


In this experiment  we assumed $\kappa=0$ (see (\ref{eq:prior})), because the training  points are located uniformly  along the full cycle of $t$, and thus there is no concentrated direction. 
We used an anisotropic exponential kernel of the form
\begin{align}
    & K((\bm{a}_{i}, s_i), (\bm{a}_{j}, s_j)) 
    \qquad i,j = 1\ldots 420,   
    \label{eq:gait_kernel}
    \\
    & = \sigma^2 \exp\left( -\frac{|| \bm{a}_{i} -\bm{a}_{j}||^2 }{2l^2} - \frac{(s_i-s_j)^2}{2g^2} \right),
    \nn 
\end{align}
Here $\bm{a}_{i} \in \mathbb{R}^3$ contais the three joint angles and $s_i \in \{ 0, \pm 5, \pm 10 \} $ indicates the surface gradient. 

At test time, we are given 20 measurements of the three joint angles, 
  with each measurement performed at a different point in the gait cycle of humans running at zero surface gradients.
  \cref{fig:gait_data} shows the  results, 
  showing a  high accuracy in the prediction of the gait cycle percentage. For sample traces and histograms see~\cref{fig:gait_traces_hist} in~\cref{app:gait_experiment}.

\subsection{The effect of Bridging}
In the experiments  presented above we did not observe a substantial increase in the MH acceptance rate when using Bridging.
However, this might change when using other kernel functions or datasets. We leave further inquires on the effect of Bridging in 
these models for future work. 

\section{CONCLUSIONS AND OUTLOOK}
\label{sec:conclusions}
In this work we explored a general model for 
Bayesian non-parametric regression of circular variables,
for which we explored efficient MCMC techniques  for learning and inference. Relevant future work includes exploring the augmentation from~\cref{sec:sampling} 
in statistical physics XY-models for which current approaches are not efficient~\citep{manon2016irreversible} (e.g. glassy regimes or 3D lattices) and exploring parameter learning using score matching~\citep{vertes2016learning}.

\subsubsection*{Acknowledgements}
AP is supported by the Israel Science Foundation (grant No. 1138/23).




\newpage 
 \bibliography{references} 




\clearpage
\newpage 

\section*{Checklist}

 \begin{enumerate}
 \item For all models and algorithms presented, check if you include:
 \begin{enumerate}
   \item A clear description of the mathematical setting, assumptions, algorithm, and/or model. [Yes]
   \item An analysis of the properties and complexity (time, space, sample size) of any algorithm. [Yes]
   \item (Optional) Anonymized source code, with specification of all dependencies, including external libraries. [Not Applicable]
 \end{enumerate}

 \item For any theoretical claim, check if you include:
 \begin{enumerate}
   \item Statements of the full set of assumptions of all theoretical results. [Yes]
   \item Complete proofs of all theoretical results. [Yes]
   \item Clear explanations of any assumptions. [Yes]     
 \end{enumerate}

 \item For all figures and tables that present empirical results, check if you include:
 \begin{enumerate}
   \item The code, data, and instructions needed to reproduce the main experimental results (either in the supplemental material or as a URL). [Yes]
   \item All the training details (e.g., data splits, hyperparameters, how they were chosen). [Yes]
         \item A clear definition of the specific measure or statistics and error bars (e.g., with respect to the random seed after running experiments multiple times). [Yes]
         \item A description of the computing infrastructure used. (e.g., type of GPUs, internal cluster, or cloud provider). [Yes]
 \end{enumerate}

 \item If you are using existing assets (e.g., code, data, models) or curating/releasing new assets, check if you include:
 \begin{enumerate}
   \item Citations of the creator If your work uses existing assets. [Yes]
   \item The license information of the assets, if applicable. [Not Applicable]
   \item New assets either in the supplemental material or as a URL, if applicable. [Not Applicable]
   \item Information about consent from data providers/curators. [Not Applicable]
   \item Discussion of sensible content if applicable, e.g., personally identifiable information or offensive content. [Not Applicable]
 \end{enumerate}

 \item If you used crowdsourcing or conducted research with human subjects, check if you include:
 \begin{enumerate}
   \item The full text of instructions given to participants and screenshots. [Not Applicable]
   \item Descriptions of potential participant risks, with links to Institutional Review Board (IRB) approvals if applicable. [Not Applicable]
   \item The estimated hourly wage paid to participants and the total amount spent on participant compensation. [Not Applicable]
 \end{enumerate}

 \end{enumerate}

\clearpage
\appendix

\onecolumn

\setcounter{figure}{0}    
\renewcommand\thefigure{S\arabic{figure}}

\aistatstitle{
Supplementary Materials}

\section{Useful definitions}
\label{app:definitions}
Given a set of angles $\phi_i$, define 
  \begin{align}
     z& = \frac{1}{n}\sum_{i=1}^n e^{i \phi_i}
     \\
     &\equiv R \exp(i\gamma) \quad R>0, \gamma \in [0, 2\pi ]
     \\
     \text{Circular mean} &\equiv \gamma
     \\
     \text{Circular variance} &\equiv 1-R
 \end{align}

\subsection{Relative Effective Sample Size (RESS)}
\label{app:RESS}
To evaluate the efficiency of the sampled Markov chains we used
Relative Effective Sample Size~\citep{liu2001monte},\footnote{For the estimation we used the \texttt{arviz} python package~\citep{kumar2019arviz}.}
defined as
\begin{equation}
    \text{RESS} = \frac{1}{1 + 2\sum_{l=1}^{\infty} \rho_l},
\label{eq:RESS}
\end{equation}
where $\rho_l$ is the autocorrelation at lag $l$.

\subsection{Circular Continuous Ranked Probability Score (CRPS)}
\label{app:crps}
The circular Continuous Ranked Probability Score (CRPS)~\citep{grimit2006continuous} is a scoring rule used to evaluate the accuracy of probabilistic forecasts by measuring the distance between the predictive distribution and the observed outcome. For circular variables, such as wind directions, CRPS is defined as:

\[
\text{CRPS}(F, \xi) = \int_0^{2\pi} \left( F(y) - \mathbf{1}(y - \xi) \right)^2 dy,
\]

where:
\begin{itemize}
    \item \( F(y) \) is the cumulative distribution function (CDF) of the predictive distribution,
    \item \( \xi \) is the holdout observation,
    \item \( \mathbf{1}(y - \xi) \) is the Heaviside step function.
\end{itemize}

In the case of circular variables, the CRPS can be simplified using the circular distance \( d(\alpha, \beta) = 1 - \cos(\alpha - \beta) \) as:

\[
\text{CRPS}(F, \xi) = \mathbb{E}\left[d(\theta, \xi)\right] - \frac{1}{2} \mathbb{E}\left[d(\theta, \theta')\right],
\]

where \( \theta \) and \( \theta' \) are independent samples from the predictive distribution \( F \).

This formulation compares the predictive distribution with the observed value and penalizes large spreads in the forecast distribution. Lower CRPS values indicate better predictive performance.


\section{The Bridging algorithm and its vMQP extension }
We first review the derivation of the Bridging algorithm of~\cite{murray2006mcmc},
and then present 
an efficient extension to von Mises Quasi-Process models.

\label{app:bridging}
\subsection{The Bridging algorithm} Since the acceptance rate (\ref{eq:exact_MH}) is known to be maximal~\citep{peskun1973optimum,tierney1998note}, one can expect to increase the acceptance rate (\ref{eq:double_MH}) 
by using a better approximation than (\ref{eq:ratio_normalizing}) to the ratio of normalizing constants.  
The Bridging algorithm~\citep{murray2006mcmc} achieves
this by using ideas from Annealed Importance Sampling (AIS)~\citep{neal2001annealed}.
The idea is to further augment (\ref{eq:augmented}) with $K$ additional variables $\bxi_k$. For this, let us introduce
$K$ distributions that interpolate 
between~$f(\cdot|w)$ 
and $f(\cdot|w')$, 
\begin{align}
 p_k(\bxi_k|w,w') & 
\varpropto     f_k(\bxi_k|w,w') \equiv 
    f(\bxi_k|w)^{\beta_k}
    f(\bxi_k|w')^{1-\beta_k}
    \qquad k=1,2\ldots K
    \label{eq_ap:pk}
\end{align}
where we defined 
\begin{align}
\beta_k = \frac{k}{K+1}  \qquad k=1,2\ldots K
\end{align}
Let us now introduce $K$ transition operators 
$T_k(\bxi_k'|\bxi_k, w,w')$ satisfying detailed balance w.r.t.~the distributions in (\ref{eq_ap:pk}), 
\begin{equation}
    p_k(\bxi_k|w,w')T_k(\bxi_k'|\bxi_k, w,w') = p_k(\bxi_k'|w,w')T_k(\bxi_k|\bxi_k', w,w')\,.
\label{eq_ap:detailed_balance}
\end{equation}
Note now that since (\ref{eq_ap:pk}) enjoys the symmetry
\begin{align}
p_k(\bxi|w,w')   = p_{K+1-k}(\bxi|w',w), 
\end{align}
we can assume that 
\begin{align}
T_k(\bxi'|\bxi, w,w') = T_{K+1-k}(\bxi'|\bxi, w',w)
\label{eq_ap:symmetry_T}
\end{align}
Using these transition operators, 
we augment the distribution (\ref{eq:conditional_w}) as 
\begin{align}
p(w,w',\bxi_0, \ldots, \bxi_K | \bvphi)  &= p(w|\bvphi) q(w'|w)
\frac{
f(\bxi_0|w')}
{Z[w']} \prod_{k=1}^K T_k(\bxi_k|\bxi_{k-1}, w,w'),
\label{eq_ap:augmented_bridge}
\end{align}
and we denoted $\bxi_0 = \bxi$. 
In order to sample from (\ref{eq_ap:augmented_bridge}), we now iterate over
\begin{enumerate}
    \item 
    Sample $w' \sim q(w'|w)$ and then $\bxi_0 \sim 
    f(\bxi_0|w')/Z[w']$. 
    \item Sample 
    \begin{align}
        \bxi_k \sim T_k(\bxi_k|\bxi_{k-1}, w,w') 
        \qquad k = 1\ldots K
        \label{eq:sample_xi_k}
    \end{align}
    \item Propose to exchange $w \leftrightarrow w'$ and 
    $\{\bxi_0, \bxi_1 \ldots \bxi_K \} \leftrightarrow \{\bxi_K, \bxi_{K-1} \ldots \bxi_0 \}$, and  accept with Metropolis-Hastings probability
\end{enumerate}
    \begin{equation}
    \min\left(1, \frac{ p(w',w,\bxi_K \ldots \bxi_0| \bvphi) }
    {p(w,w',\bxi_0 \ldots \bxi_K| \bvphi) } =\frac{p(w') q(w|w') f(\bvphi|w')}
    {p(w) q(w'|w)  f(\bvphi|w)} 
    \prod_{k=0}^K \frac{f_{k+1}(\bxi_k|w,w')}
    {f_{k}(\bxi_k|w,w')} 
    \right). 
    \label{eq_ap:bridging_MH}
    \end{equation}

The  product expression in the rhs follows directly from (\ref{eq_ap:detailed_balance})-(\ref{eq_ap:augmented_bridge}), and 
we extended the definition~(\ref{eq_ap:pk}) to include $f_0(\cdot|w,w') = f(\cdot|w')$ and    
$f_{K+1}(\cdot|w, w') = f(\cdot|w)$. 
Comparing (\ref{eq:double_MH}) with (\ref{eq_ap:bridging_MH}), 
we see that the approximation (\ref{eq:ratio_normalizing}) has been replaced by 
\begin{align}
    \frac{Z[w]}{Z[w']} \simeq 
        \prod_{k=0}^K \frac{f_{k+1}(\bxi_k|w,w')}
    {f_{k}(\bxi_k|w,w')}, 
\qquad \qquad \bxi_k \sim     f_k(\bxi_k|w,w'),
    \label{eq_ap:AIC}
\end{align}
which is an AIS estimate~\citep{neal2001annealed} of the ratio of normalization constants. 

\subsection{Efficient Bridging for von Mises Quasi-Processes via double augmentation}
\label{subsec:efficient_bridge}
Back to our von Mises model, the $K$ intermediate distributions (\ref{eq_ap:pk}) are
\begin{align}
p_k(\bxi_k |w,w')  \varpropto f_k(\bxi_k|w,w') = \exp \left\{  
-\frac12 
\sum_{i,j=1}^d 
M_{k,ij}\cos(\xi_{k, i}-\xi_{k,j}) 
+
\alpha_{k,c} 
\sum_{i=1}^d
\cos(\xi_i)
+ \alpha_{k,s} 
\sum_{i=1}^d
\sin(\xi_i)
\right\}    
\label{eq_ap:von_Mises_k}
\end{align}
for  $k =1 \ldots K$, 
where we defined 
\begin{align}
\bm{M}_k &= \beta_k \bm{M}_{w}  + \beta_{K+1-k} \bm{M}_{w'}. 
\label{eq_ap:Mk}
\\
\alpha_{k,c} &= 
\beta_k \kappa \cos(\nu)
+
(1-\beta_k) \kappa' \cos(\nu')
\\
\alpha_{k,s}  &=
\beta_k \kappa \sin(\nu)
+
(1-\beta_k) \kappa' \sin(\nu')
\end{align}
Transition operators $T_k$ satisfying detailed balance (\ref{eq_ap:detailed_balance}) w.r.t.~these distributions can be obtained for each $k$ using the method of~\cref{sec:sampling}, 
\begin{align}
    T_k(\bxi_k|\bxi_{k-1},w,w') = 
    p(\bxi_k|\z_k,w,w')     p(\z_k|\bxi_{k-1},w,w')  \qquad k =1 \ldots K,
\end{align}
where $p(\z_k|\bxi_{k-1},w,w')$ 
and $p(\bxi_k|\z_k,w,w')$ are defined as in (\ref{eq:z_given_phi}) and (\ref{eq:product_vMises})-(\ref{eq:kappas}) respectively, 
using a matrix $\A_k \in \mathbb{R}^{d  \times d}$ from a Cholesky decomposition of $\bm{I}_d \lambda_k- \bm{M}_k$ with an appropriate $\lambda_k$. 
Note that we treat the variables $\z_k$ as part of the randomness of the transition. 
However, this approach would require to perform 
$K$ Cholesky decompositions for each proposal $w'$, 
hurting  the computational advantage we seek to gain 
for large $K$. 

Instead, we show here that we can do well 
by reusing the two decompositions, 
\begin{align}
    \A^\top_{w}  \A_{w} &= \bm{I}_{d} \lambda_{w} - \bm{M}_{w} ,
    \\
    \A^\top_{w'}  \A_{w'} &= \bm{I}_{d} \lambda_{w'} - \bm{M}_{w'} ,
\end{align}
which were performed when sampling 
fictitious data points from $f(\bxi|w)$ and $f(\bxi|w')$ in the previous and present Doubly MH steps, respectively. 
Recall that $\lambda_{w}, \lambda_{w'}$ were chosen the make the right-hand sides  positive definite. 
The idea is to use two Gaussian augmentations instead of one,  
to separately eliminate each term of~$\M_k$ in  (\ref{eq_ap:Mk}). 
More concretely, for each $k$ let us define 
\begin{align}
    p_k(\y_1|\bxi) = & {\cal N}
    \left( \y_1; \beta_k^{1/2} \A_w \cos(\bxi),  \bm{I}_{d} \right) 
&
    p_k(\y_2|\bxi) = & {\cal N}
    \left( \y_2; \beta_k^{1/2}  \A_w \sin(\bxi),  \bm{I}_{d} \right) 
    \label{eq_ap:pk_y}
\\
    p_k(\y_3|\bxi) = & {\cal N}
    \left( \y_3; \beta^{1/2}_{K+1-k}  \A_{w'} \cos(\bxi),  \bm{I}_{d} \right) 
&
    p_k(\y_4|\bxi) =  & {\cal N}
    \left(\y_4; \beta^{1/2}_{K+1-k} \A_{w'} \sin(\bxi),  \bm{I}_{d} \right) 
    \label{eq_ap:pk_yp}
\end{align}
Let us augment (\ref{eq_ap:von_Mises_k}) now as 
\begin{align}
p_k(\bxi, \y |w,w') 
    = &  p_k(\bxi|w,w')     p_k(\y_1|\bxi)     p_k(\y_2|\bxi) 
        p_k(\y_3|\bxi)     p_k(\y_4|\bxi)
        \label{eq_ap:pk_xi_y}
        \\
\varpropto &       
\exp \left\{ \bk_{c} \cdot \cos(\bxi) 
    + 
\bk_{s} \cdot \sin(\bxi) 
    -\frac12 \y^\top \y     \right\} , 
\end{align}
where we defined $\y = [\y_1, \y_2, \y_3, \y_4]$ and 
\begin{align}
\bk_{c} = \beta_k^{1/2}   \y_1^{\top} \A_w + \beta_{K+1-k}^{1/2}   \y_3^{\top} \A_{w'}
+ \alpha_{k,c} {\bm{1}_d}
\qquad 
\bk_{s} = \beta_k^{1/2}   \y_2^{\top} \A_w + \beta_{K+1-k}^{1/2}   \y_4^{\top} \A_{w'} \,.
+ \alpha_{k,s} {\bm{1}_d}
\end{align}
where ${\bm{1}_d}$ is a $d$-dimensional vector of $1$'s.

We choose as the  operator $T_k$ that satisfies detailed balance 
w.r.t.~(\ref{eq_ap:von_Mises_k}) the marginal w.r.t.~$\y$ of the  Gibbs sampling  operator that leaves  invariant the augmentation (\ref{eq_ap:pk_xi_y}). 
Thus to sample (\ref{eq:sample_xi_k}), 
we perform Gibbs steps in (\ref{eq_ap:pk_xi_y}), 
\begin{align}
\y & \sim p_k(\y|\bxi_{k-1},w,w'),
\label{eq:bridge_sample1}
\\
\bxi_k & \sim   p_k(\bxi_k|\y,w,w'),     
\quad k=1\ldots K
\label{eq:bridge_sample2}
\end{align}
and discard $\y$, 
where $p_k(\y|\bxi_{k-1},w,w')$ is the product of the four 
densities in (\ref{eq_ap:pk_y})-(\ref{eq_ap:pk_yp}) and
\begin{equation}
    p_k(\bxi_k|\y,w,w') \varpropto \prod_{i=1}^d \exp(\alpha_i \cos(\xi_{k,i} -\gamma_i) ) \,, 
\label{eq:product_vMises2}
\end{equation}
where 
\begin{align}
    \alpha_i &= \sqrt{\kappa_{c,i}^2 + \kappa_{s,i}^2} 
\qquad \quad 
    \tan (\gamma_i)  = \frac{\kappa_{s,i}}{\kappa_{c,i}}  \,.
\end{align}

\section{Details of the experiments} 
The experiments were performed on a desktop computer with 
an Intel Core i5-13400F CPU at 2.90GHz. 

\label{app:experiments}
\subsection{Experiment: Wind directions in Germany} 
\label{app:wind_experiment}
We run the sampler for 200K iterations, discarding the first 20K as burn-in. 

\begin{figure}
\begin{center}
\fbox{
\includegraphics[width=.95\linewidth]{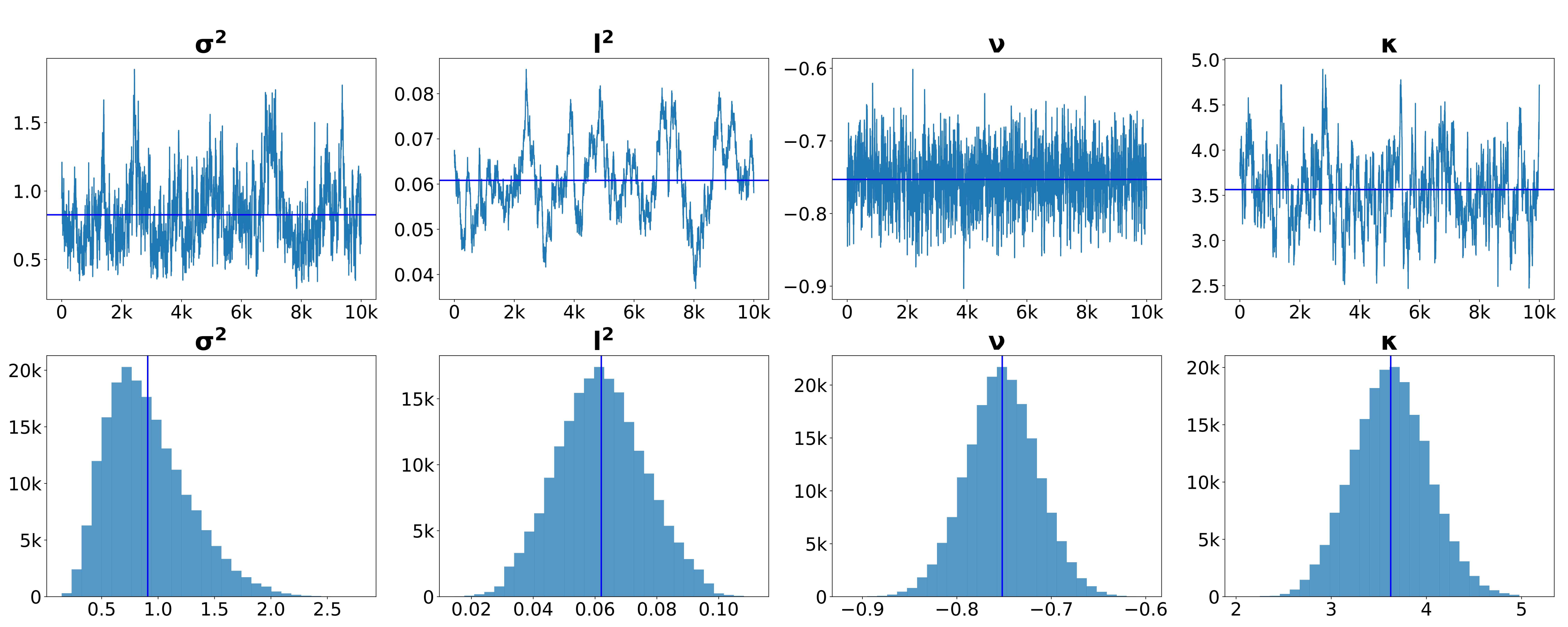}
}
\caption{Sample traces and histograms of the parameters of the vMQP model (\ref{eq:prior}) with kernel (\ref{eq:gaussian_kernel}) in the experiment of wind directions prediction from~\cref{sec:wind_predictions}.}
\label{fig:wind_par_traces}
\end{center}
\end{figure}

\begin{figure}
\begin{center}
\fbox{
\includegraphics[width=.95\linewidth]{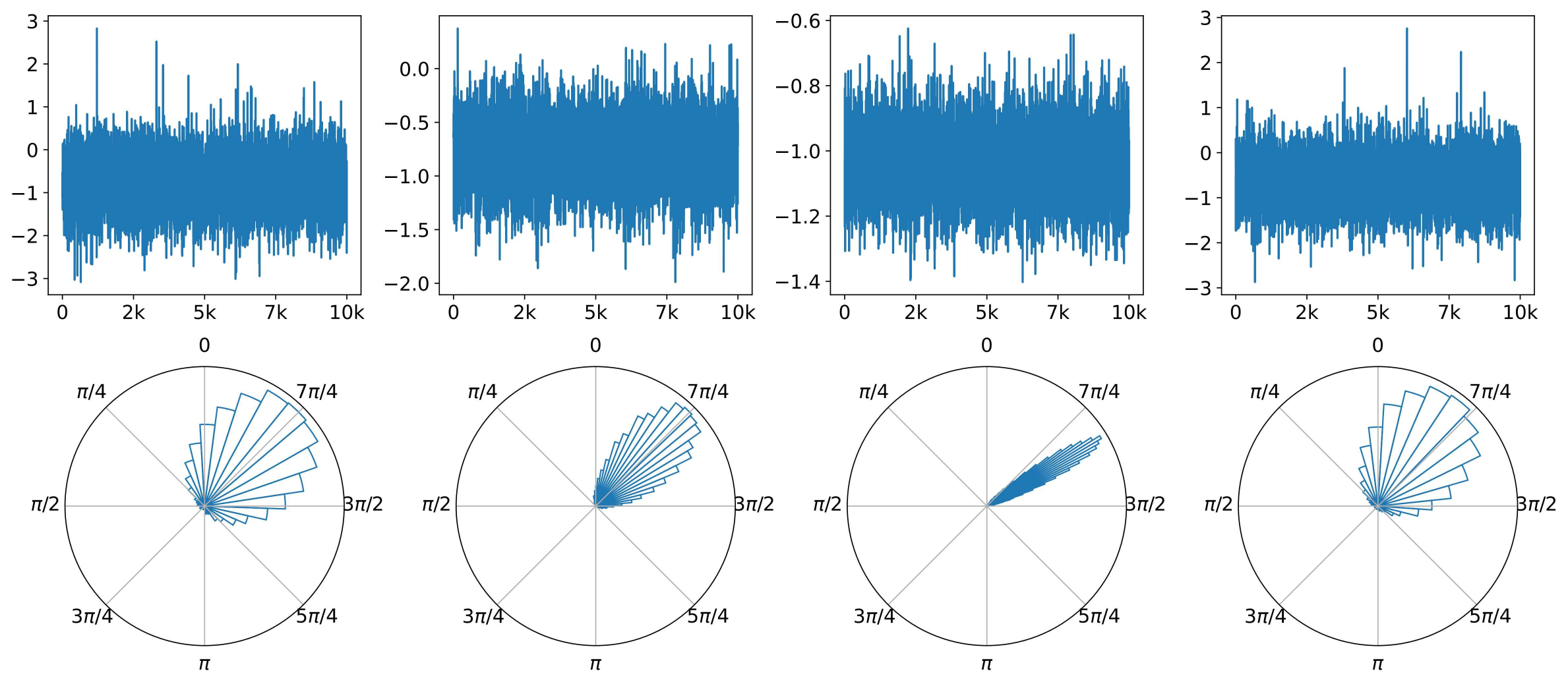}
}
\caption{Sample traces and circular histograms of the predicted distribution of wind directions from~\cref{sec:wind_predictions} at four test locations.}
\label{fig:wind_angles_traces}    
\end{center}
\end{figure}

\begin{itemize}
    \item Parameter priors: 
    \begin{align}
        \sigma^2 & \sim  {\cal N}(\sigma^2; 0, 1) \qquad \text{ truncated to }\sigma^2 >0
        \\
        l^2 & \sim {\cal N}(l^2; 0, 1) \qquad \text{ truncated to }l^2 >0
        \\
        \kappa & \sim {\cal N}(\kappa; 0, 1) \qquad \text{ truncated to }\kappa >0  
        \\
        \nu & \sim \text{Uniform} [-\pi, +\pi]
    \end{align}
    \item MH Proposals: random walk on all the parameters, with variances adjusted for reasonable acceptance rates.
\end{itemize}

\subsection{Experiment: Prediction of percentage of running cycle  from joints angles}
\label{app:gait_experiment}
\begin{itemize}
    \item Parameter priors: 
    \begin{align}
        \sigma^2 & \sim  {\cal N}(\sigma^2; 0, 1) \qquad \text{ truncated to }\sigma^2 >0
        \\
        l^2 & \sim {\cal N}(l^2; 0, 1) \qquad \text{ truncated to }l^2 >0
        \\
        g^2 & \sim {\cal N}(g^2; 0, 1) \qquad \text{ truncated to }g^2 >0  
    \end{align}
    \item MH Proposals: random walk on all the parameters, with variances adjusted for reasonable acceptance rates.
\end{itemize}

\begin{figure}[t!]
\begin{center}
\fbox{
\includegraphics[width=0.95\textwidth]{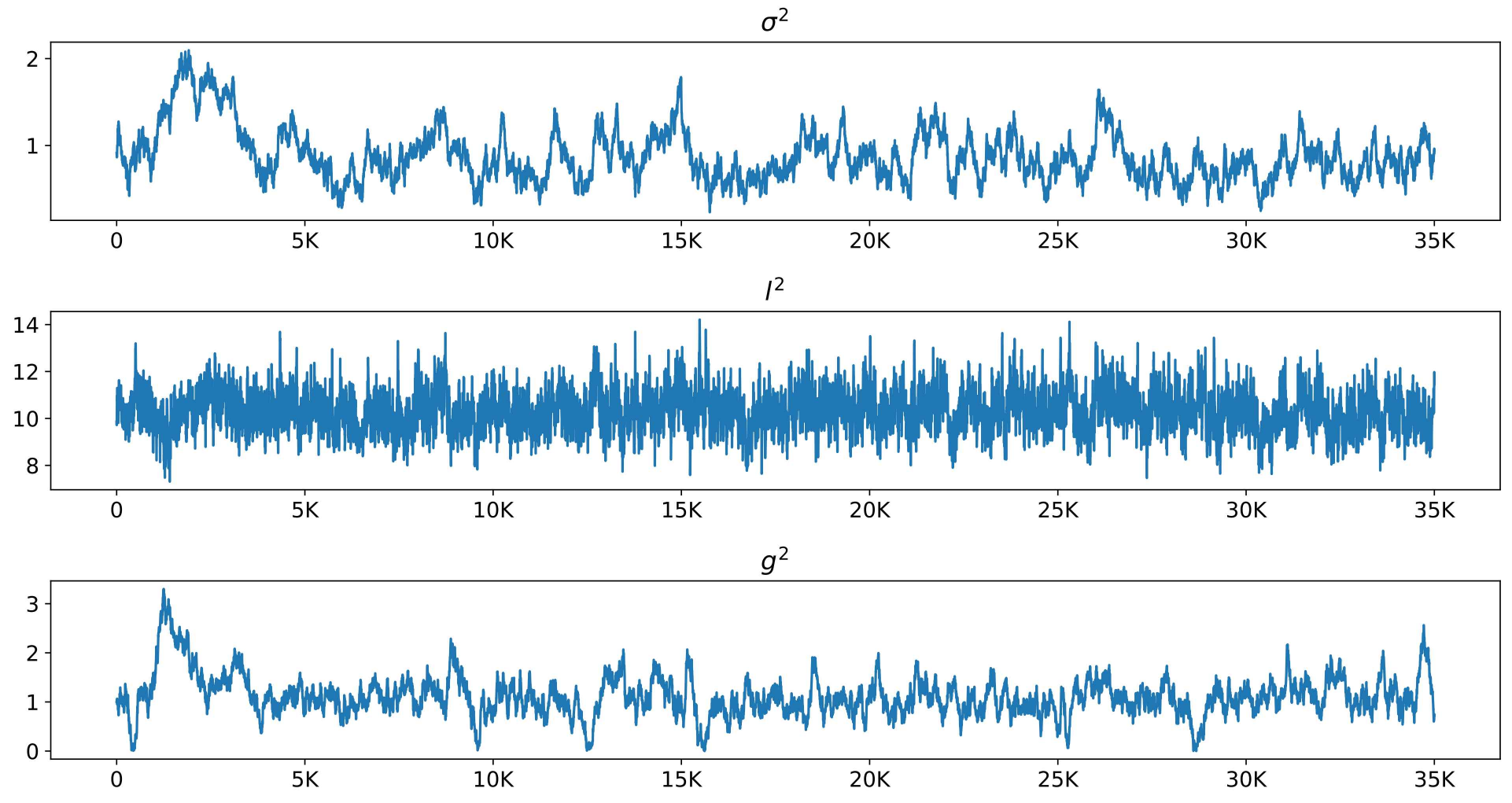}
}
\vskip .1cm
\fbox{
\includegraphics[width=0.95\textwidth]{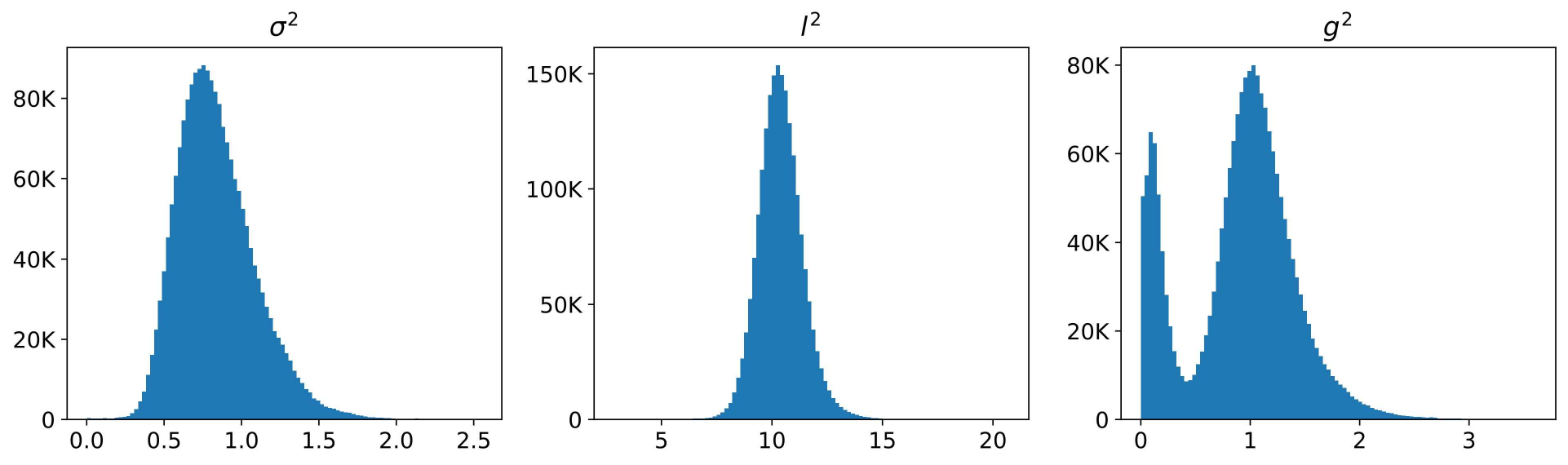}
}
\caption{Sample traces and histograms of the parameters of the kernel (\ref{eq:gait_kernel}) in the experiment of gait running cycle prediction from~\cref{sec:percentage_prediction}. Note the bimodality of the third parameter, correctly captured by our Bayesian approach. } 
\label{fig:gait_traces_hist}
\end{center}
\end{figure}

We run the sampler for 600K iterations, discarding the first 20K as burn-in.  

\begin{itemize}
    \item Parameter priors: positively truncated Gaussians for $\sigma^2, l^2, g^2$. 
\end{itemize}

\textbf{Experimental protocol used to collect the data.} 
In the analysis, we used running data from~\cite{shkedy2022parametric}  who tested 16 healthy adults, with no lower-limb injuries or impairments: 9 males and 7 females (age: 24.56 ± 3.16 years, [Range 18–28 years]; height: 1.73 ± 0.09 m, [range: 1.55–1.86 m]; mass: 68.01 ± 13.98 kg, [range: 45.0–88.7 kg]). All participants run on an instrumented split-belt treadmill at a speed of 2.25 m/s and at five different surface gradients (-10\%, -5\%, 0\%, +5\%, +10\%). Motion and force data were collected at each surface gradient using a minimum of 7 gait cycles, with an average of 16 cycles per condition. Using this data for each participant, the average joint kinematics and kinetics were calculated. 
Motion data were collected using a motion capture system at a frequency of 120 Hz to capture the positions of 22 reflective markers attached to each participant’s pelvis and right leg (modification of the Calibration Anatomical System Technique (CAST) marker set~\citep{cappozzo1995position}). The raw data of marker positions was low-pass filtered (Butterworth second order forward and backward passes) with a cut-off frequency of 10 Hz.

Using the Visual 3D software~\citep{v3d},
three lower-limb joint angles were obtained.
these were ankle-, knee-, and hip-joint angles in sagittal plane (side view), as defined in~\cref{fig:gait} (Right). 
The  angles were calculated for the right leg only (since both legs behaved symmetrically). The joint angles were normalized in time as percentages of one stride cycle using spline interpolation. All calculated joint angles for each participant were averaged across all the measured gait cycles.

\end{document}

%% file: packages.tex
\usepackage{graphicx}
\usepackage{siunitx}
\graphicspath{ {./images/} }

\usepackage{array}

\usepackage{amsfonts}       
\usepackage{mathtools}
\usepackage{enumitem}
\usepackage{bm}
\usepackage{amsmath}
\usepackage{amsthm}
\usepackage{amssymb}
\usepackage{pifont}

\usepackage{graphicx}
\usepackage{caption}

\usepackage{amsfonts}
\newcommand{\pmat}{\begin{pmatrix}}
\newcommand{\pman}{\end{pmatrix}}

\usepackage{authblk}

%% file: macros.tex
%
%



\newcommand{\eq}{\begin{equation*}}
\newcommand{\en}{\end{equation*}}
\newcommand{\eqa}{\begin{eqnarray*}}
\newcommand{\ena}{\end{eqnarray*}}
\newcommand{\eqn}{\begin{equation}}
\newcommand{\enn}{\end{equation}}
\newcommand{\be}{\begin{equation}}
\newcommand{\ee}{\end{equation}}
\newcommand{\eqan}{\begin{eqnarray}}
\newcommand{\enan}{\end{eqnarray}}

\newcommand{\nn}{\nonumber}


\newcommand{\A}{ {\bf A} }

\newcommand{\M}{ {\bf M} }

\newcommand{\f}{\textbf{f}}
\newcommand{\y}{ {\bf y} }
\newcommand{\x}{ {\bf x} }
\newcommand{\z}{ {\bf z} }

\newcommand{\bt}{\bm{\theta}}
\newcommand{\bphi}{\bm{\phi}}
\newcommand{\bvphi}{\bm{\varphi}}

\newcommand{\bk}{\bm{\kappa}}

\newcommand{\bxi}{\bm{\xi}}



 


\newcommand{\argmax}{\mathop{\mathrm{argmax}}}


%% file: transductive.tex
\begin{figure}[t!]
\begin{center}
\includegraphics[width=.48\textwidth]{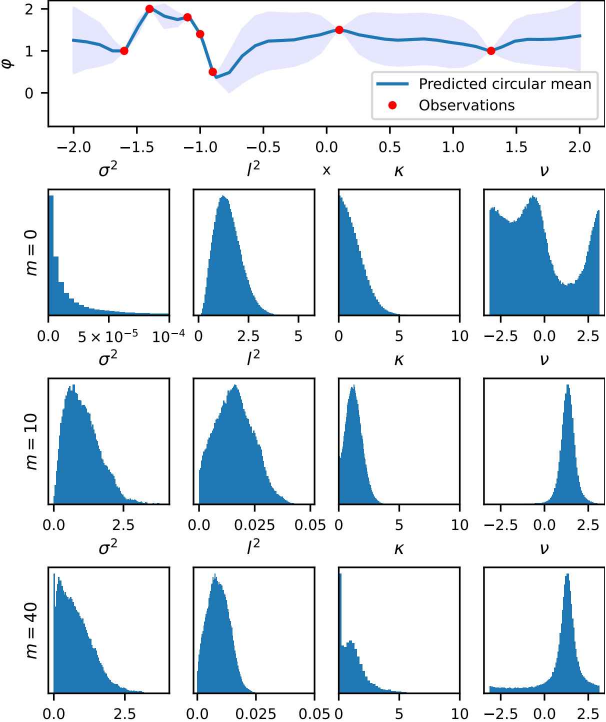}
 \caption{ {\bf Transductive learning in action.} 
Given seven angular observations on the $x$ axis, 
we show histograms of posterior samples of the parameters of a vMQP  model (\ref{eq:prior}) with kernel $K(x_i,x_j) = \sigma^2 
 \exp(-(x_i-x_j)^2/2l^2)$, for different numbers $m$ of uniformly located predictive locations. Transductive learning manifest itself
 in the changes of these distributions as a function of the predictive locations. 
 Note the shrinking of $l^2$ as $m$ grows and the multimodality of $\nu$ captured by the Bayesian approach.  
  The confidence interval in the top panel is proportional to the circular variance for $m=40$.  
}
\label{fig:synth_data}
\end{center}
\end{figure}

%% file: cont_div_histograms.tex
\begin{figure}[t!]
\begin{center}
\includegraphics[width=.48\textwidth]{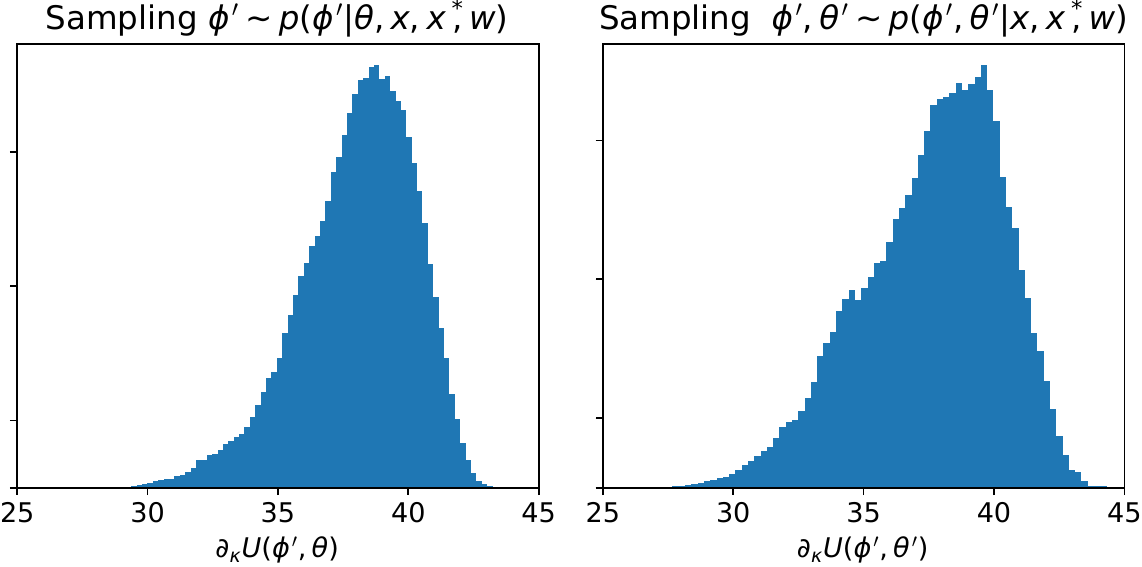}
 \caption{ {\bf Problems with maximum likelihood estimation.}  Histograms of two different random evaluations
 of $\partial_{\kappa}U(\bphi, \bt)$,  
 whose difference is required 
 to estimate the log-likelihood gradient (\ref{eq:cd_gradient})  w.r.t.~$\kappa$. The samples are from the model 
 in~\cref{fig:synth_data} with $m=40$.
 The similarity of the distributions leads to highly inaccurate gradient estimates.  
 }
\label{fig:cont_div}
\end{center}
\end{figure}

%% file: ess_lambda.tex
\begin{figure}[t!]
\begin{center}
\includegraphics[width=.48\textwidth]{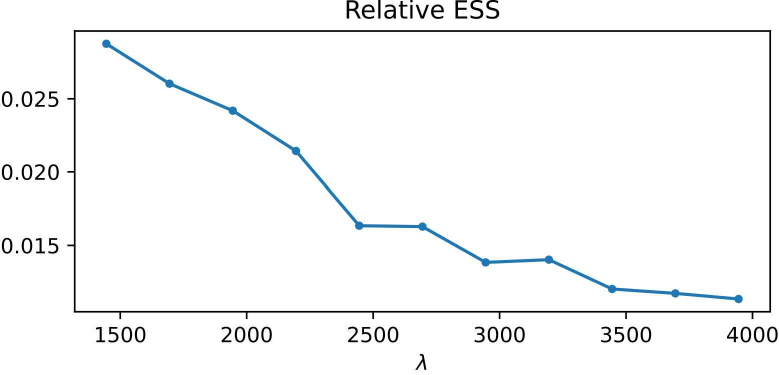}
 \caption{ {\bf Optimal $\lambda$.} Relative Effective Sample Size (ESS) (see definition in~\cref{app:RESS})
 of the log of the density~(\ref{eq:conditional_density}), computed with samples from the data presented in~\cref{fig:synth_data} with $m=10$, as a function of the parameter $\lambda$ that 
enables the Cholesky decomposition~(\ref{eq:Cholesky}). 
The plot confirms that smaller $\lambda$
should be preferred.
}
\label{fig:ess_lambda}
\end{center}
\end{figure}

%% file: sampler_performance.tex
\begin{figure}[t!]
\begin{center}
\includegraphics[width=.48\textwidth]{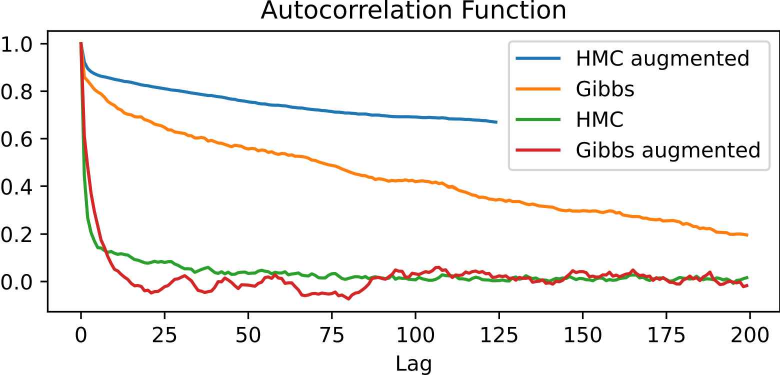}
 \caption{ {\bf Samplers comparison.} Autocorrelation function of the 
 same log-density as in~\cref{fig:ess_lambda}, 
 computed using the augmented Gibbs and Hamiltonian Monte Carlo on the non-augmented (\ref{eq:conditional_density}) and augmented
(\ref{eq:joint}) models. Evaluations  were made  between iterations such that the CPU time spent on each sample is roughly equal.
}
\label{fig:sampler_performace}
\end{center}
\end{figure}

%% file: wind_test_table.tex

\begin{table*}[t]
    \centering
    \setlength{\tabcolsep}{8pt}
    \sisetup{separate-uncertainty=true}
    \renewcommand{\arraystretch}{1.2} 
    \begin{tabular}{lS[table-format=1.3(3)]S[table-format=1.3(3)]S[table-format=1.3(3)]S[table-format=1.3(3)]}
        \toprule
        \multicolumn{1}{c}{\textbf{Model}} & \multicolumn{4}{c}{\textbf{CRPS}$\downarrow$} \\
        \cmidrule(lr){2-5}
        & {\textbf{10\% Test}} & {\textbf{20\% Test}} & {\textbf{30\% Test}} & {\textbf{40\% Test}} \\
        \midrule
        Wrapped GP        & 0.221(86) & 0.283(93) & 0.305(175) & 0.399(204) \\
        Projected GP      & 0.067(15) & 0.068(9)  & 0.063(5)   & 0.072(9)   \\
        vMQP Gaussian     & $\bf{0.057\pm 0.011}$ & 0.064(6)   & 0.066(7)   & 0.066(10)  \\
        vMQP Exponential  & 0.058(15) & $\bf{0.060\pm 0.004}$  & $\bf{0.057\pm 0.006}$  & $\bf{0.061\pm 0.008}$   \\
        \bottomrule
    \end{tabular}

    \caption{Model performance comparison using CRPS (see~\cref{app:crps}) 
    across various test set sizes. Results show {\it mean $\pm$ standard deviation} from seven random data partitions. 
    }
    \label{tab:wind}
\end{table*}

%% file: wind_directions.tex
\begin{figure*}[h] 
\begin{center}
\includegraphics[width=.97\textwidth]{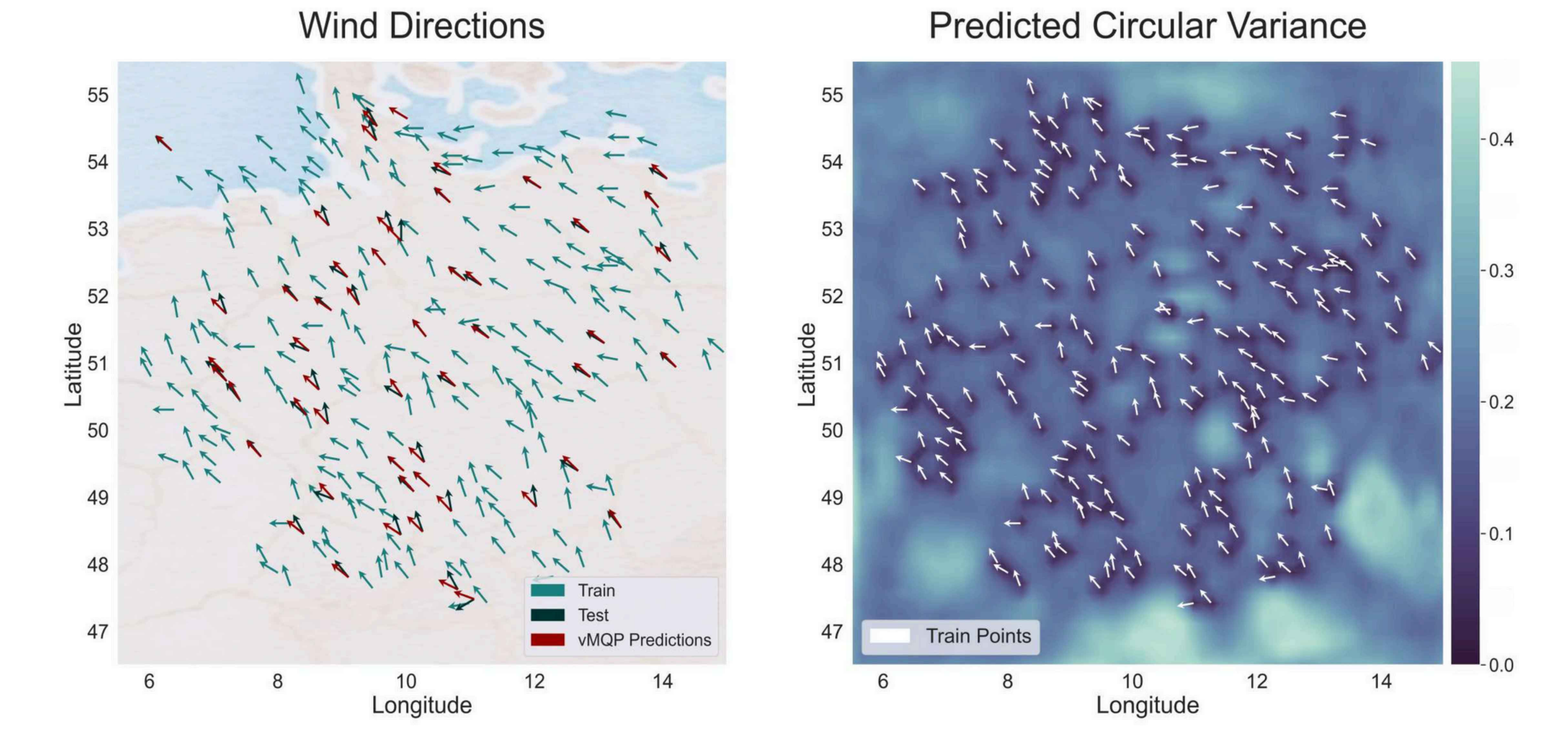}

 \caption{{\bf Left:} Wind directions in 
 260 weather stations in Germany, randomly split between 
  208 train and 52 test locations. The predicted circular means from the  vMQP model are indicated on the test locations.
{\bf Right:} Predicted circular variance over a uniform grid of 
$60 \times 60$ points. Note that the variance 
grows in regions close to train points with non-aligned directions. 
Both figures best seen in color.
}
\label{fig:wind_direction_variance}
\end{center}
\end{figure*}

%% file: gait.tex
\begin{figure*}[t]
\begin{center}
\fbox{
\includegraphics[width=0.98\textwidth]{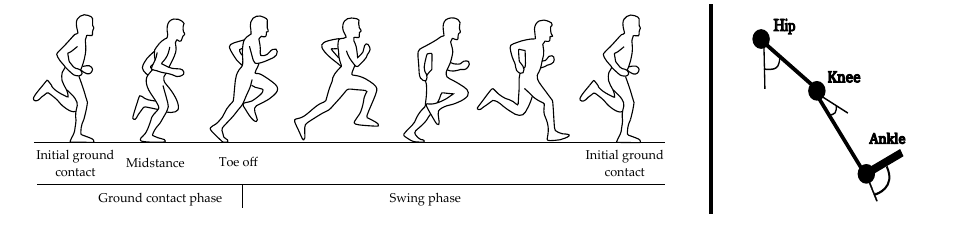}
}
\caption{{\bf Left:} Phases of the running gait cycle, defined w.r.t.~the right leg of the figure. The cycle begins and ends with the initial ground contact of the right leg. 
Image:~\cite{phd2022running}.
{\bf Right: } The three joint angles used to predict the location
in the gait cycle,  defined over the sagittal plane. 
}
\label{fig:gait}
\end{center}
\end{figure*}

%% file: joints_test.tex
\begin{figure}[h!]
\begin{center}
\includegraphics[width=.38\textwidth]{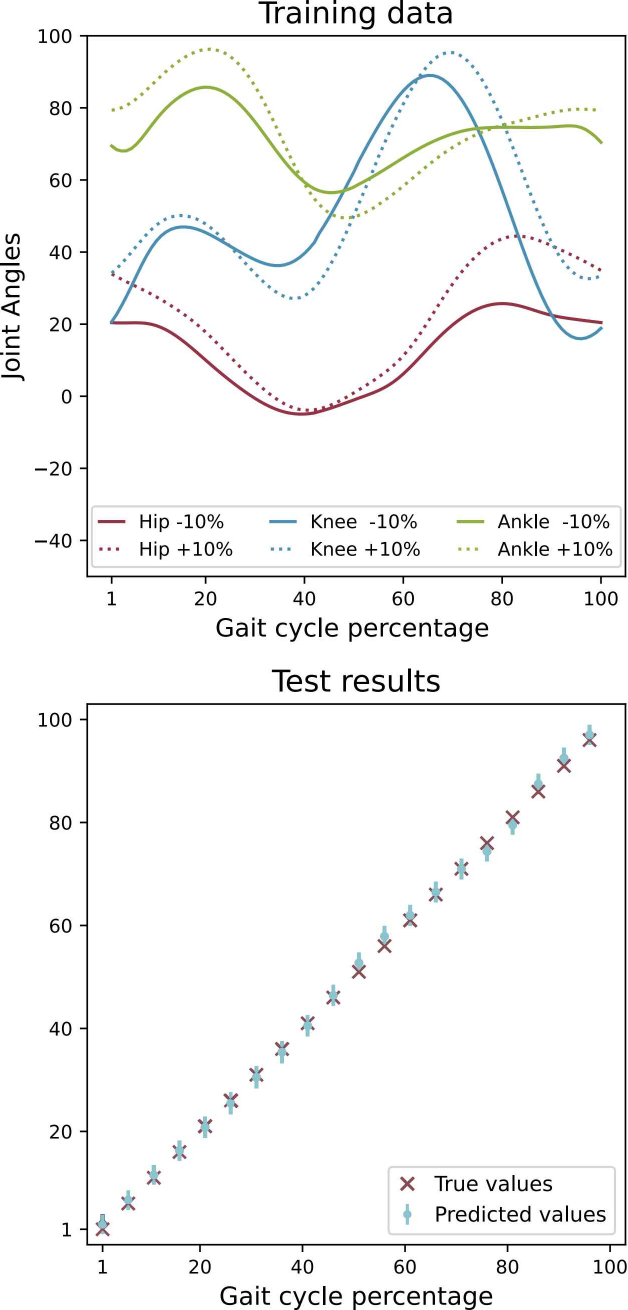}

 \caption{{\bf Predictions of gait cycle percentage}. {\it Top:} training data for $\pm 10 \%$ gradients.
 {\it Bottom:}
 Predictions for 20 data points  at zero gradient. The error bars are proportional to the circular variance. 
}
\label{fig:gait_data}
\end{center}
\end{figure}
